\documentclass[journal]{sty/IEEEtran}

\usepackage{cite}
\usepackage{color,xcolor}
\usepackage{graphicx}
\usepackage{pdfpages}
\usepackage{subfigure}
\usepackage{tabulary}
\usepackage{balance}
\usepackage{amsmath}
\usepackage{amssymb}
\usepackage{float}
\usepackage{wasysym}
\usepackage{soul}
\usepackage{booktabs} 
\usepackage{multirow}  
\usepackage{hyperref} 
\usepackage{array}  

\newcommand{\PreserveBackslash}[1]{\let\temp=\\#1\let\\=\temp}
\newcolumntype{C}[1]{>{\PreserveBackslash\centering}p{#1}}
\newcolumntype{R}[1]{>{\PreserveBackslash\raggedleft}p{#1}}
\newcolumntype{L}[1]{>{\PreserveBackslash\raggedright}p{#1}}

\usepackage{makecell,rotating,multirow,diagbox}
\graphicspath{{figures/}}
\usepackage{bm}
\usepackage{verbatim}

\usepackage{amsfonts}
\usepackage{pifont}

\usepackage{sidecap}
\usepackage{hyperref}

\usepackage{tikz}

\renewcommand{\baselinestretch}{.99} 

\begin{document}
	
	\title{Learning Purposeful Co-designs of Bio-inspired Quadrupedal Robots} 
	
	\author{Fabio Boekel$^1$, Jiatao Ding$^{2,\star}$, and Cosimo Della Santina$^1$ 
		\thanks{This work is supported by the EU project 101016970 NI.}
		\thanks{Fabio Boekel and Cosimo Della Santina are with Cognitive Robotics, Delft University of Technology, Building 34, Mekelweg 2, 2628 CD Delft, Netherlands (e-mail: F.G.M.Boekel@student.tudelft.nl, C.DellaSantina@tudelft.nl).  
			Jiatao Ding is with AI Core, Beijing Zhongguancun Academy, Beijing, 100094, China (email: dingjiatao@bza.edu.cn).
			Cosimo Della Santina is also with the Institute of Robotics and Mechatronics, German Aerospace Center (DLR), 82234 Wessling, Germany (e-mail: fabioboekel@live.com, cosimodellasantina@gmail.com).}
		\thanks{Jiatao Ding is the corresponding author.}
	}

	\markboth{xxx}%
	{Shell \MakeLowercase{\textit{et al.}}: Bare Demo of IEEEtran.cls for IEEE Journals}
	
	\maketitle
	
	\begin{abstract}
		Treating morphology and control as separate design problems has constrained the efficiency and adaptability of quadrupedal robots, while current co-design approaches are often constrained by restricted design spaces.  We overcome this limit by introducing \textit{Dogger}, a bio-inspired quadruped with modular three-segment legs, four-bar linkages and adjustable passive knee compliance. A two-stage framework combining reinforcement learning and Bayesian optimization is developed: the morphology-conditioned control policy is trained across designs and then held fixed during task-specific design selection. The resulting designs and policies deployed zero-shot reduce the cost of transport by 18.6\% and 35.7\% for inspection and payload-carrying tasks. Comparisons demonstrate that full geometry--compliance co-design outperforms fixed and restricted design scopes. Cross-task tests further reveal distinct task-specific morphologies with shared mechanical trade-offs. These results provide hardware-level evidence that expanding the physical design space alongside learned control is essential to realize the full potential of task-driven morphology–control co-design in legged robotics.
	\end{abstract}
	
	
	\IEEEpeerreviewmaketitle
	
	\section{Introduction}
	\label{introduction}

	Animals exhibit diverse forms of agile, efficient and adaptable locomotion through the close integration of morphology, neural control and behaviour. This coupling continues to inspire the design of legged robots capable of meeting different task demands \cite{stella2025synergy,seok2014design,dingego,humphreys2025learning}.
	
	Inspired by the natural muscle-skeleton system, passive elastic actuators, especially parallel elastic actuators (PEAs), are incorporated into the quadrupedal design, improving robustness, reducing energy consumption or generating highly dynamic motions \cite{buchli2006finding,sprowitz2013towards,sprowitz2018oncilla,bjelonic2023learning,arm2019spacebok,ruppert2022learning}. 
	On the other hand, learning-based controllers have enabled quadrupeds to traverse challenging terrain \cite{miki2022learning,kim2025high,lee2020learning}, locomote over deformable surfaces \cite{choi2023learning}, and perform agile or adaptive multi-skill behaviours \cite{hwangbo2019agile,han2024lifelike,kang2026agile,yang2020multi}. 
	Yet these controllers are typically developed for a specified mechanical platform, whose leg geometry and passive mechanics shape the dynamics and energy exchanges underlying locomotion. Changes to these properties may therefore require policy retraining or hardware-specific adjustment.
	Physical legged robots provide concrete examples: Badri-Spr\"owitz et al.\ developed BirdBot, which uses avian-inspired geometry and passive clutching to support energy-efficient locomotion with minimal active control \cite{badri2022birdbot}; Ruppert and Badri-Spr\"owitz matched a closed-loop central pattern generator (CPG) to passive elastic dynamics \cite{ruppert2022learning}; and Stella \textit{et al.} embedded behavioral synergies in tendon routing, pulley geometry, and joint stiffness \cite{stella2025synergy}. Together, these studies highlight the importance of coordinating mechanical structure and control, as further illustrated by the system-level energy-efficiency improvements demonstrated by RAIBO2 \cite{lee2026quadruped}. They motivate a broader design question: How should morphology and control be jointly optimized to meet different task demands? 
	
	Morphology-control co-design offers a promising way to address this question. Evolutionary approaches search mechanical design and gait or controller parameters \cite{digumarti2014concurrent}, while trajectory-optimization and control-aware methods incorporate design variables into motion optimization \cite{ha2017joint,de2021control,dinev2022versatile}. Reinforcement learning (RL) approaches have explored concurrent learning of agent design and behaviour, including frameworks that couple online and offline learning \cite{schaff2019jointly,Ha2018designrl,won2019learning,luck2020data,chen2023c}. 
	Other approaches combine learning with trajectory optimization or evolutionary search \cite{grandesso2023reinforcement,gupta2021embodied}, or couple evolutionary design search with optimal control \cite{fadini2024co}. Meta-RL and design-conditioned policies further support the evaluation of candidate morphologies by reusing learned control across designs \cite{belmonte2022meta,bjelonic2023learning}.
	Collectively, these approaches have broadened the available optimization strategies. However, an algorithmic formulation alone does not determine which mechanical variables can be changed in a real robot, how readily candidate designs can be fabricated and exchanged, or whether predicted improvements persist after hardware deployment.
	
	Recent studies have begun to address these physical challenges. Song \textit{et al.} designed 
	task-oriented quadrupedal leg mechanisms, introducing a differential three-segment design for payload and efficiency \cite{song2025codesign}. Most directly related to this work, Bjelonic \textit{et al.} combined a design-conditioned RL policy with Bayesian optimization (BO) to co-design passive knee elasticity and locomotion control on ANYmal \cite{bjelonic2023learning}. Their study demonstrated learning-based physical co-design with a mechanical search focused on passive knee elasticity. Design optimization used a teacher policy, whereas a distilled student policy was deployed on hardware. 
	These advances demonstrate the practical value of co-design, while leaving open how different mechanical design freedoms should be combined for a given task. In particular, optimizing passive compliance within a prescribed leg geometry does not establish whether jointly varying geometry and compliance yields additional benefits, or how these benefits change across tasks. 
	Systematically investigating these questions on hardware calls for a platform that supports controlled variation of geometry and compliance, together with a control framework for evaluating candidate designs and deploying the selected configurations.
	
	In this work, we jointly expand the mechanical design space and the learning framework through developing a modular hardware platform and proposing a morphology-aware design strategy. By aligning an expressive physical substrate with advanced co-design algorithms, we provide a systematic real-world demonstration that morphology–control co-design yields substantial gains in quadrupedal locomotion under specific task requirements. This overarching contribution unfolds through three key technical advances:
	\begin{itemize}
		\item \textbf{Modular quadruped \textit{Dogger}.} We introduce a modular quadruped for systematic morphological variation. Its bio-inspired three-segment legs and parallel-elastic knee elements enable rapid changes to geometry and passive compliance while retaining the same trunk, actuators, sensing and onboard computing, providing an expanded, physically realizable design space for co-design.
		\item \textbf{Morphology-aware learning framework.} We develop a two-stage co-design framework that combines RL and BO. A morphology-conditioned policy is trained across a distribution of leg geometries and compliance parameters, then held fixed while BO selects task-specific designs. The same policy evaluates candidate designs and controls the selected hardware without retraining or distillation.
		\item \textbf{Experimental validation of task-dependent design benefits.} We demonstrate zero-shot sim-to-real transfer of optimized designs and policies. Relative to a nominal spring-disengaged configuration, the selected hardware reduces system-level cost of transport by 18.6\% for inspection and 35.7\% for payload transport, providing quantitative evidence of real-world efficiency gains from expanded morphology-control co-design. Comparison studies show that joint geometry–compliance optimization outperforms fixed and restricted design scopes, while cross-task evaluations reveal task-specific performance trade-offs. 
	\end{itemize}
	
	\section{Methodology}
	This section details the co-design methodology for the bio-inspired quadruped, focusing on the three primary components: (1) the bio-inspired leg design, (2) the learning-based co-design algorithm, and (3) the sim-to-real strategy.
	
	\subsection{Bio-inspired Leg Design}
\begin{figure}
		\centering
		\includegraphics[width=0.999\linewidth]{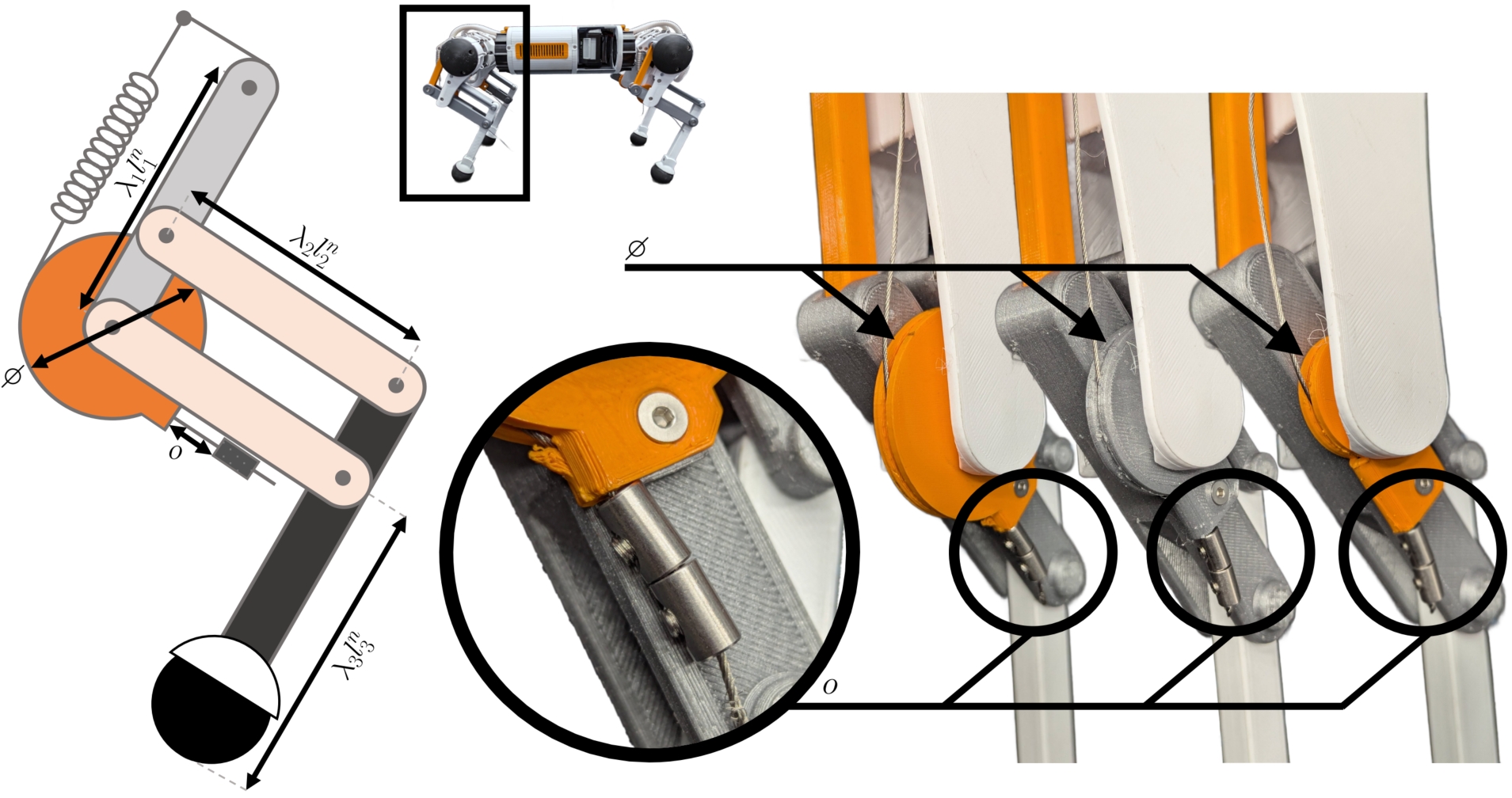}
		\vspace{-4mm}
		\caption{\textbf{Design of the bio-inspired leg.} The lengths of the leg’s links are scaled from their nominal values ($l_1^n$, $l_2^n$, $l_3^n$) using scaling factors ($\lambda_1$, $\lambda_2$, $\lambda_3$). The spring wheel’s diameter ($\diameter$) adjusts the effective rotational stiffness, while the spring offset ($o$) specifies the engagement angle. The middle inset shows the clamps used to set $o$. The right panel shows three spring-wheel configurations, with $\diameter$ decreasing from left to right. The wire extending upward is connected to the spring, which is not shown here.}
		\label{fig:designspace}
	\end{figure}

	As can be seen in Fig.~\ref{fig:designspace}, the quadruped has four identical legs, each of which consists of two main components: a three-segment mechanism with four-bar linkage kinematics and a parallel spring at the knee joint. This bio-inspired configuration expands the mechanical design space compared to a traditional two-segment non-elastic leg.
	
	The leg morphology is determined by the three link lengths (thigh, shank and foot) and the mechanical properties of the knee spring system (rotational stiffness and spring engagement angle). Given three scaling factors: thigh ($\lambda_1$), shank ($\lambda_2$), and foot ($\lambda_3$), each link length is determined by its respective scaling factor and nominal length (denoted  by $l_1^n$, $l_2^n$, $l_3^n$, respectively). The rotational joint stiffness and the engagement angle are controlled by the wheel diameter ($\diameter$) and spring offset ($o$). Therefore, a specific design $\boldsymbol{d}$ in the design space $\mathcal{D}$ is parametrized as $\boldsymbol{d} = [\lambda_1, \lambda_2, \lambda_3, \diameter, o]^\text{T}$.
	
	\subsubsection{Three-segment leg mechanism}
	Inspired by animal anatomy, we design the leg with three segments, with a parallel linkage implemented for the third segment. The leg structure is fully 3D printed and designed to exclude any electronics and belts. Modifications to $\lambda_1,\lambda_2,\lambda_3$ can be made by reprinting individual parts. The entire leg structure can be removed in one piece without the need to detach a single motor. These modular design features enable rapid and cost-effective testing of the co-design pipeline.
	
	The scale parameters $\lambda_1,\lambda_2,\lambda_3$ range from $0.70$ to $1.30$, with the bounds determined by the minimum and maximum sizes that can fit in the robot body. Reducing the lower limit was difficult because the parallel linkage joint in the thigh would interfere with the motor's output axis, while increasing the upper limit would result in a collision between the front and back legs of the robot. The nominal scaling factor is centred within this range at $1.00$.
	
	\subsubsection{Knee PEA mechanism}
	We further integrate a PEA into the knee joint to emulate the elasticity found in biological systems. Inspired by the wheel-wire mechanism in \cite{bjelonic2023learning, dingego, liu2018switchable}, we incorporate a spring in the thigh joint, to which a wire is connected. The wire passes over a wheel positioned at the knee joint, with both the wheel and the wire attached to the shank. In this design, the diameter $\diameter$ of the wheel influences the rotational stiffness of the joint. Since the wheel is 3D printed, it can be easily swapped, enabling faster implementation, as there is no need to purchase a new spring each time the robot is optimized with a different spring stiffness. The mechanism with different wheels can be seen on the right side of Fig.~\ref{fig:designspace}.
	
	Furthermore, this design allows $o$ to be adjusted by altering the position of the clamps on the wire, effectively modifying the leg angle at which the spring activates. As can be seen in the middle inset of Fig.~\ref{fig:designspace}, the position of the clamp can be changed by screwing or unscrewing the screw on the clamp and moving it along the wire.
	
	For the PEA, we choose a spring with a stiffness of $c_s=4180$ N/m, with $\diameter$ ranging from $0$ to $50$\,mm and $o$ from $0$ to $10$\,mm. With these ranges, we can achieve a rotational joint stiffness between $0.03-2.62$ Nm/rad and an initial torque on the knee joint in the standing position between $0-5.62$ Nm. Across these parameter ranges, spring deformation remains below the limit at which plastic deformation occurs. 
	
	
	\subsection{Learning-based Morphology-control Co-design}
\begin{figure*}
		\centering
		\includegraphics[width=\textwidth]{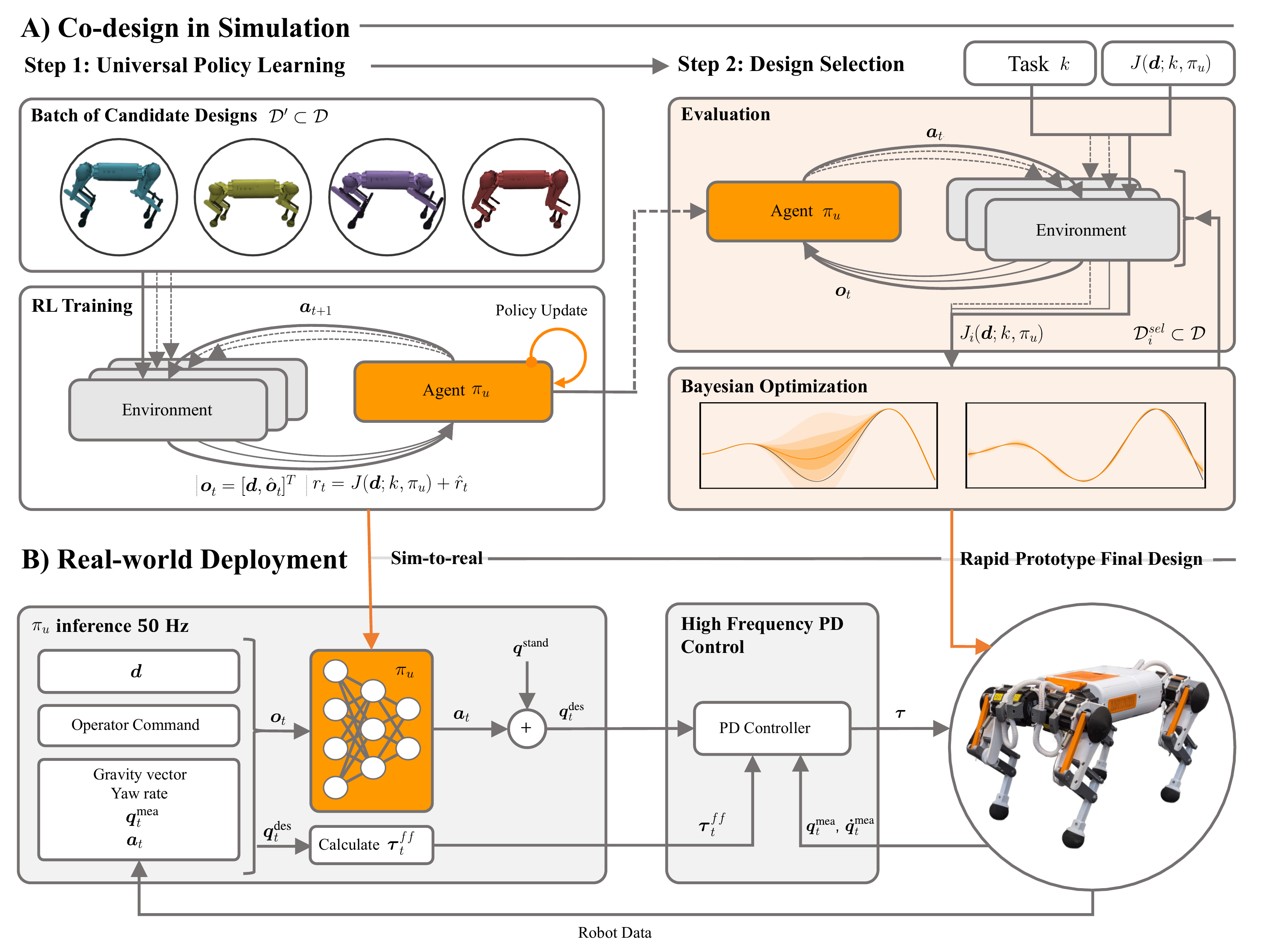}
		\vspace{-8mm}
		\caption{\textbf{Two-stage optimization for morphology-control co-design.} In \textit{Step 1}, a universal control policy $\pi_u$ is trained in simulation across a broad design space $\mathcal{D}$ using RL. In \textit{Step 2}, a BO-based selector identifies the optimal design $d^*_k$ for a specific task $k$ by evaluating energy performance. For real-world deployment, the policy maps observations $\boldsymbol{o}_t$ to actions $\boldsymbol{a}_{t}$ while incorporating the feedforward spring torque $\boldsymbol{\tau}_{t}^\text{ff}$, which helps to bridge the sim-to-real gap.}
		\label{fig:algorithmoverview}
	\end{figure*}

	Our learning-based co-design pipeline uses a two-step process to optimize both the morphology and the controller, given a task (Fig.~\ref{fig:algorithmoverview}). This process consists of 1) universal policy training using RL and 2) design selection through BO. In the first step, we train a universal policy $\pi_u$ that is adapted to robot designs ($\boldsymbol{d}$) distributed across the design space $\mathcal{D}$. In the second step, we use $\pi_u$ to control the different designs for a specific task $k$ and select the best design with the lowest cost. After the two-step optimization, we obtain the task-specific design $d^*_k$ and the universal controller $\pi_u$ used to control it.
	
	\subsubsection{RL-based universal policy training}
	Formally, RL is formulated as a Markov decision process (MDP). At each timestep $t$, the agent observes its state $\boldsymbol{s}_t$, in our case, a partial observation $\boldsymbol{o}_t$, and takes an action $\boldsymbol{a}_t$. This interaction results in a transition to a new state $\boldsymbol{s}_{t+1}$, and the agent receives a reward $r_t$ that evaluates the quality of its action. Using the new state $\boldsymbol{s}_{t+1}$ and its policy $\pi(\boldsymbol{a}_{t+1} | \boldsymbol{s}_{t+1})$, the agent takes the next action $\boldsymbol{a}_{t+1}$. The goal of the RL algorithm is to optimize the policy to maximize the cumulative rewards obtained over time.

	The observation space, action space, rewards, and termination follow.
	
	\textbf{Observation space:} The observation includes both the design parameters (\(\boldsymbol{d}\)) and the remaining portion of the observation (\(\hat{\boldsymbol{o}}_t^\prime\)). This portion \(\hat{\boldsymbol{o}}_t^\prime\) consists of the commanded forward (\textit{x}) and sideward ({\textit{y}}) velocities, the commanded yaw rate, the gravity vector of the robot’s body, the actual yaw rate, the joint positions $\boldsymbol{q}_t^{mea}$, and the previous actions $\boldsymbol{a}_{t}$. Together, these elements form the current observation \(\boldsymbol{o}_t^\prime = [\boldsymbol{d}, \hat{\boldsymbol{o}}_t^\prime]^\text{T}\).
	
	\textbf{Action space:}
	The action space consists of the desired joint angles. $\boldsymbol{a}_{t}$ is added to the predefined standing pose of the robot \(\boldsymbol{q}^{\text{stand}}\), resulting in the desired joint angles \(\boldsymbol{q}_{t}^{\text{des}}\).
	
	The commands generated by the policy are executed by actuators, which operate using the internal proportional-derivative (PD) controller. The PD controller generates the torques to track \(\boldsymbol{q}_{t}^{\text{des}}\). See Fig.~\ref{fig:algorithmoverview}B for an overview of how the actions are processed.
	
	
	\textbf{Rewards:}
	The reward function \(r_t\) combines task-specific and regularization terms. The task-specific reward focuses on velocity tracking, whereas the regularization terms promote motion smoothness, base stability and energy efficiency. In our work, the reward structure is designed to avoid biasing the policy toward specific designs. For example, we exclude reward terms such as body height, which could bias a specific design. Note that the reward also incorporates the cost function \(J(\boldsymbol{d}; k, \pi_u)\) for the design selection (as discussed in the second step), ensuring alignment between the training and design selection objectives.
	
	
	\textbf{Termination:} Each episode terminates when the robot flips upside down, reaches joint limits or exceeds the maximum episode length. 
	
	Our implementation uses a single fixed $\pi_u$. Unlike \cite{belmonte2022meta}, which uses a meta-RL approach to adapt policies across different designs, our simpler approach achieves robust performance across the considered designs, thereby simplifying the training pipeline. While \cite{bjelonic2023learning} distinguishes between design-aware and deployment policies, we use a single policy informed by design parameters, ensuring that the deployment policy is also aware of the robot design. In addition, in contrast to \cite{bjelonic2023learning, belmonte2022meta}, we do not use motion primitives for the foot trajectories as part of the reward, enhancing the freedom to explore the best motion for each design.
	
	\subsubsection{BO-based design selection}
	\label{sec:designselection}
	In the second step, we select the best design ${d}^*_{k}$ by minimizing a cost function $J(\boldsymbol{d};k,\pi_u)$ for a specific task $k$.
	
	The task \(k\) defines the environment scenario and the task requirements. It can include a specific velocity command, a particular type of terrain, disturbance levels, or variations in the robot, such as a version with a payload. Given a task, we aim to improve energy efficiency by minimizing the cost of transport $CoT$, defined as
	\begin{equation}
		J(\boldsymbol{d};k,\pi_u): CoT = \frac{E_\text{total}}{\Delta s} = \frac{\int_{0}^{T} P_\text{total}(t)dt}{\Delta s},
		\label{eq:CoT}
	\end{equation}
	where the total energy usage $E_\text{total}$ is calculated from the integration of the total power of the system $P_\text{total}(t)$ over time, \(\Delta s\) is the total distance, and $T$ is the total time.
	
	The cost function $J(\boldsymbol{d};k,\pi_u)$ above enables fair comparisons across different velocity commands and designs with varying velocity tracking performance, which, however, is not differentiable with respect to $\boldsymbol{d}$. Therefore, we utilize a derivative-free optimization method. In this work, we chose BO over an evolutionary algorithm because of its sample efficiency. Following the standard BO procedure, this design selection process takes place over multiple iterations. At each iteration \(i\), the BO algorithm receives the cost function scores of a batch of designs \(\mathcal{D}^{sel}_{i}\) from the previous iteration. Based on these scores, it selects a new batch of designs \(\mathcal{D}^{sel}_{i+1}\). During each episode, the trained policy \(\pi_u\) performs inference to control the different designs in \(\mathcal{D}^{sel}_{i+1}\). At the end of the episode, the scores of $J(\boldsymbol{d};k,\pi_u)$ for these designs are returned to the BO algorithm. This process is repeated for a set number of iterations to find the best design \({d}^*_{k}\).
	
	
	\subsection{Sim-to-real Transfer}
	To ensure the zero-shot transfer of purposeful co-designs from simulation to the physical world, we employ the following sim-to-real strategies: domain randomization, noise injection, feedforward compensation for the PEA and a comprehensive energy model.
	
	\subsubsection{Domain randomization}
	Domain randomization is applied by adjusting parameters such as body weight and inertia, PD gains, and contact friction. These ranges are selected based on real-robot testing to encompass the variability observed in real-world environments. Note that domain randomization is not applied during the second design selection step to ensure that performance differences between candidate morphologies are attributable to the design itself rather than stochastic environment variations.
	
	\subsubsection{Disturbances and noise injection}
	During $\pi_u$ training, random velocity perturbations are introduced to simulate unexpected disturbances that the robot may encounter in real environments. In addition, random noise is added to all observation values except for the design parameters. Noise is also added to the spring offset and resampled every 100 steps.
	
	In the second step, the same disturbances and noise are applied. To ensure a fair comparison, we share the random seed across all tested designs, ensuring that each design experiences identical noise and disturbances during evaluation.
	
	\subsubsection{PEA simulation}
	\label{sec:simtorealpea}
	The additional knee torque from the parallel spring is estimated with
	\begin{equation}
		\tau_{i}(q_i(t), \diameter, c_s) = \max \left( \frac{1}{4}\diameter^2 c_s (q_i(t)) + \frac{1}{2} \diameter c_s o, \, 0 \right),
		\label{eq:springff}
	\end{equation}
	\noindent where $\tau_{i}(q_i(t), \diameter, c_s)$ is the torque that is produced by the spring on joint $i$, which depends on the joint angle $q_i(t)$, spring setpoint $o$ and the spring stiffness $c_s$.
	
	We compute a feedforward torque command $\boldsymbol{\tau}_{t}^\text{ff}$ using \eqref{eq:springff} to compensate for spring torque on the knee joint, allowing the policy to optimize the gait and utilize the spring effectively rather than compensate for spring-induced deviations.
	
	\subsubsection{Comprehensive energy model}
	To ensure that the purposeful co-designs in simulation reflect the real efficiency gains of hardware, we employ a comprehensive energy model that accounts for both the mechanical work of the bio-inspired legs and the static electrical overhead of the platform. The total power consumption (\(P_\text{total}(t)\)) is estimated as the sum of the power consumed by the twelve motors and the power consumption of the remaining electronics, such as the NVIDIA Jetson computer. 
	Denoting the total motor power as \(P_\text{mech}(t)\) and other electronics as \(P_\text{system}\), we have
	\begin{equation}
		P_\text{total}(t) = P_\text{mech}(t) + P_\text{system}.
		\label{eq:power}
	\end{equation}
	
	Measurements have shown that \(P_\text{system}\) remains relatively constant during operation and is therefore time-independent. Inspired by \cite{bellegarda2020robust}, we estimate \(P_\text{mech}(t)\)  by
	\begin{equation}
		\begin{split}
			P_\text{mech}(t) &= \sum_{i=1}^n P_{\text{mech},i}(\tau_i(t), \dot{q}_i(t)) \\
			&= \sum_{i=1}^n \left[ \left(\frac{\tau_i(t)}{K_T}\right)^2 R + \frac{K_v \tau_i(t) \dot{q}_i(t)}{K_T} \right],
		\end{split}
		\label{eq:motorpower}
	\end{equation}
	where \(\tau_i(t)\) is the torque applied by the motor \(i\), \(\dot{q}_i(t)\) is the rotational speed, \(R\) is the motor's winding resistance, \(K_T\) is the torque constant, and \(K_v\) is the velocity constant. This equation accounts for both winding power dissipation, proportional to \(\tau_i(t)^2 R_i\), and rotational power, \(\tau_i(t) \dot{q}_i(t)\).
	
	
	\section{Results}
	We evaluate physical co-design through complementary simulation and hardware experiments on the \textit{Dogger} platform. The main results focus on task-specific design selection, cross-task performance, contributions of geometry and passive compliance, and robust locomotion on rough terrain. 
	
	\subsection{Design and Evaluation Settings}
	
	We train the universal control policy using proximal policy optimization (PPO)~\cite{schulman2017proximal} in thousands of parallel environments, each assigned a different design sampled from the morphology space. In our implementation, up to 8,192 environments are used for flat-terrain training within the Brax framework~\cite{freeman2021brax}, while more complex terrain scenarios are trained using NVIDIA Isaac Lab~\cite{mittal2025isaac} with 4,096 environments. Training requires approximately 15 minutes to 2 hours on a desktop GPU (NVIDIA RTX 4070), depending on the task complexity. The resulting controller operates at a frequency of $50\,$Hz.
	

	
	For design selection, we use the tree-structured Parzen estimator (TPE) solver \cite{ozaki2022multiobjective} provided by the Optuna library \cite{optuna_2019}. The process begins with a warm-up phase in which 100 randomly sampled designs are evaluated. Each design is evaluated for 6000 environment steps in the inspection task and 15000 environment steps in the payload transport task. 
	
	Detailed settings for the RL reward, domain randomization, and BO are attached in the appendix. Hardware devices are also specified in the appendix.

	\subsection{Task-Driven Co-design for Energy-Efficient Locomotion}
\begin{figure}
		\centering
		
		\begin{minipage}{0.48\textwidth}
			\centering
			\includegraphics[width=0.99\textwidth]{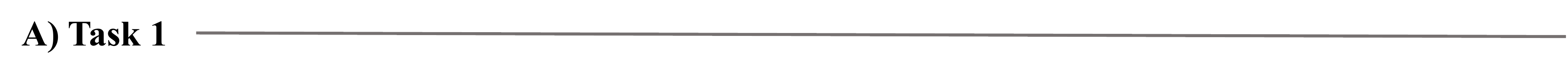}
		\end{minipage}
		
		\begin{minipage}{0.48\textwidth}
			\centering
			\includegraphics[width=0.99\textwidth]{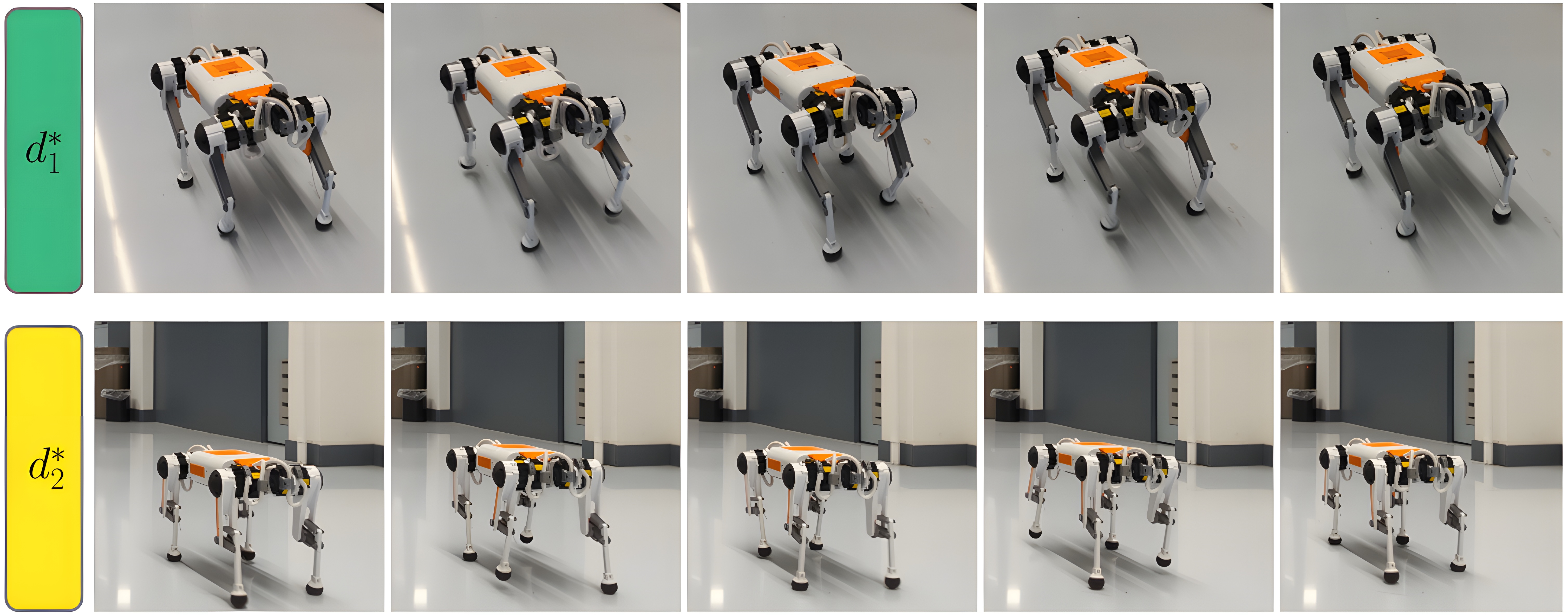}
		\end{minipage}
		\begin{minipage}{0.48\textwidth}
			\centering
			\includegraphics[width=0.99\textwidth]{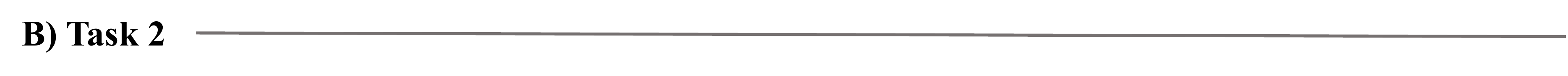}
		\end{minipage}
		
		\begin{minipage}{0.48\textwidth}
			\centering
			\includegraphics[width=0.99\textwidth]{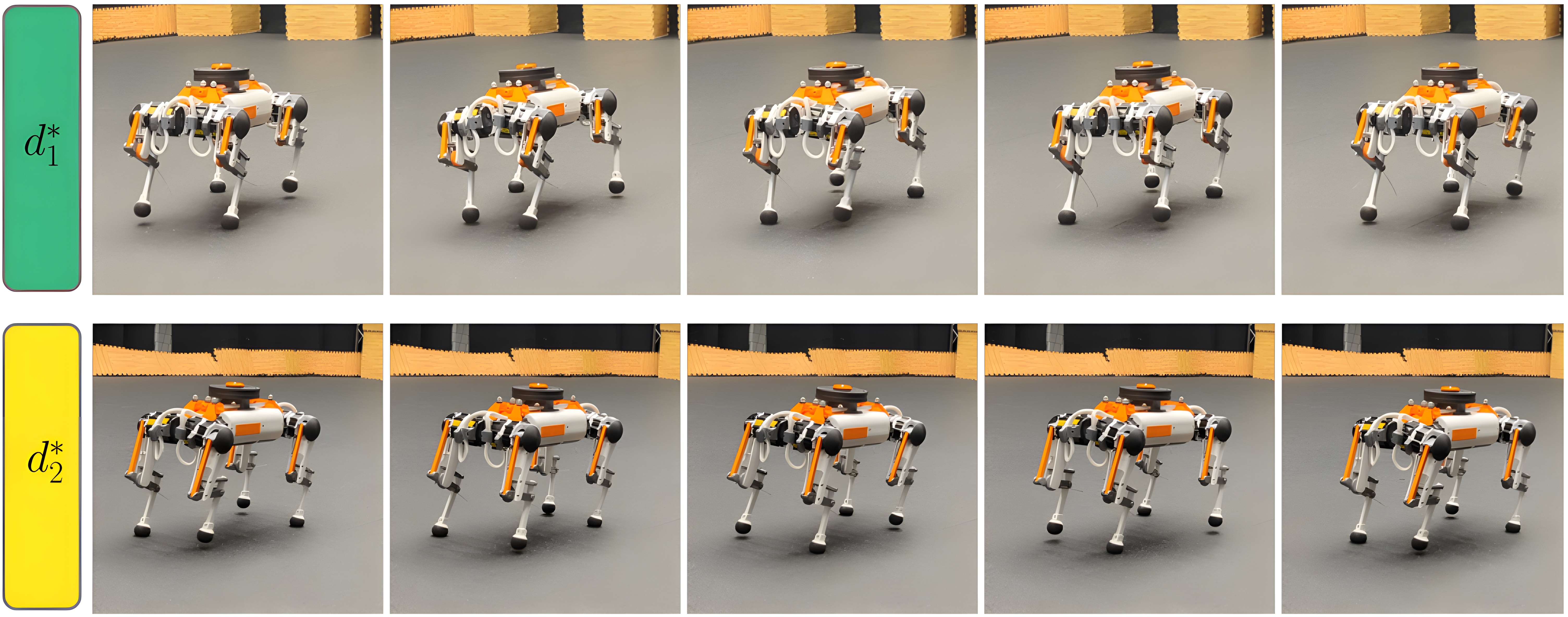}
		\end{minipage}
		
		\caption{\textbf{Snapshots from indoor experiments.} For task 1, the robot walks forward at $0.5$\,m/s. For task 2, the robot walks sideways to the left at $0.25$\,m/s. Consecutive frames are separated by $0.12$\,s. All four selected designs were tested on hardware; only a subset is shown because of space constraints.}
		\label{fig:experiment2grid}
	\end{figure}

	We asked whether different task objectives select distinct energy-efficient physical configurations and whether their predicted advantages transfer to hardware. We considered two design optimization tasks: task 1 for efficient inspection and task 2 for efficient payload transport, and evaluated each optimized design in both simulation and hardware experiments (see Fig.~\ref{fig:experiment2grid} and \href{https://youtu.be/klQ1qpdMpb8}{Movie~1}). The resulting designs are denoted by $d^*_1$ for task 1 and $d^*_2$ for task 2. These designs were compared with two nominal reference designs whose link lengths were set to be their mid-range values: nominal design 1, $d^n_1$, with the springs disengaged, and nominal design 2, $d^n_2$, with the springs set to be their mid-range values. 
	
	\begin{figure}[!t]
		\centering
		
		\includegraphics[width=\linewidth]
		{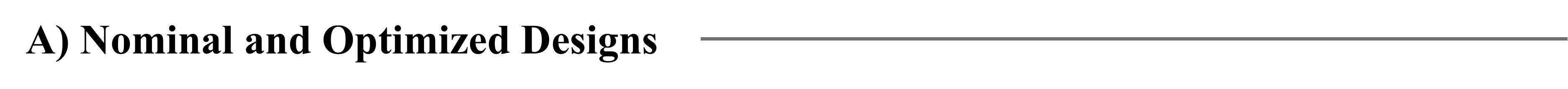}\par
		\begin{minipage}[c]{0.49\linewidth}
			\centering
			\includegraphics[width=\linewidth]
			{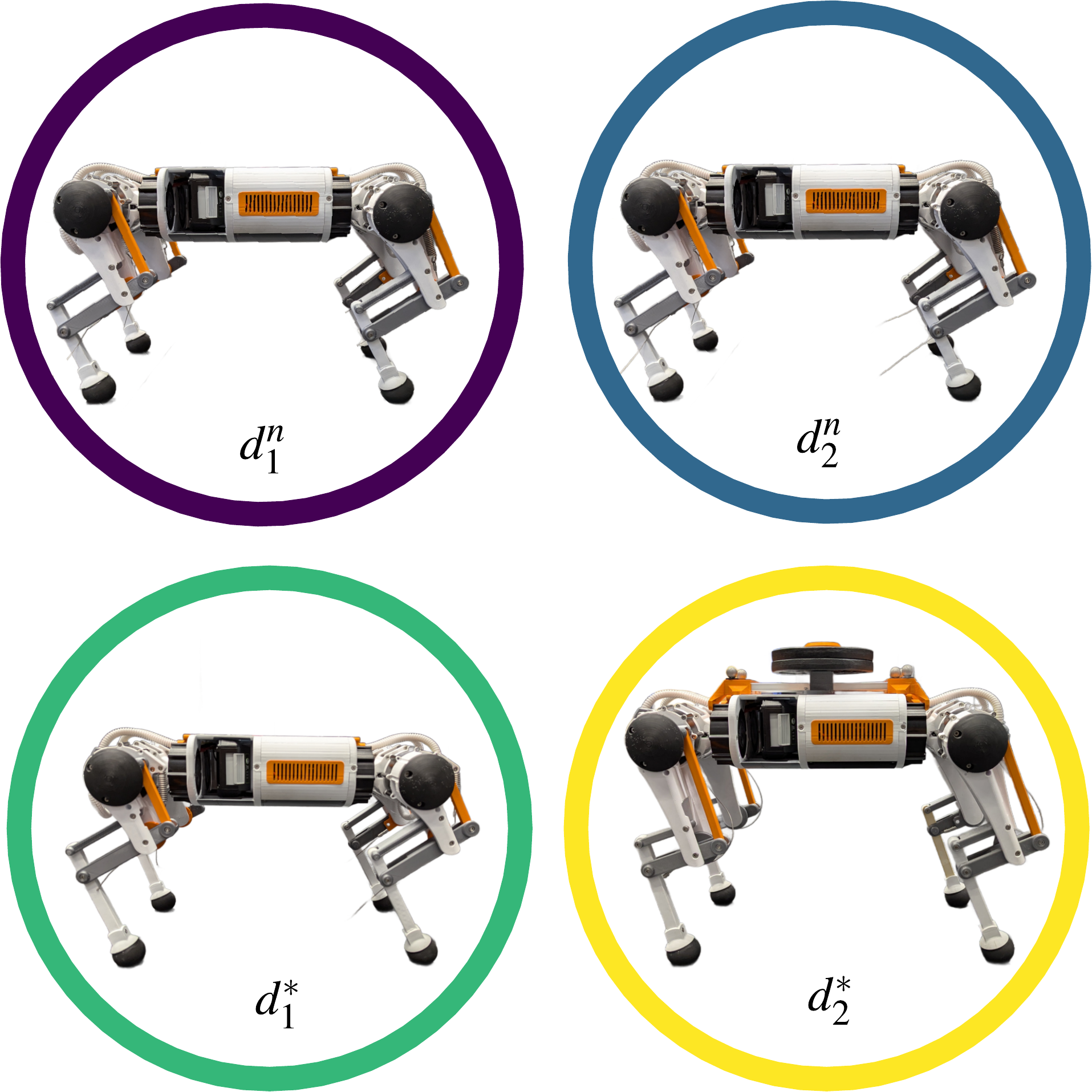}
		\end{minipage}\hfill%
		\begin{minipage}[c]{0.49\linewidth}
			\centering
			\includegraphics[width=\linewidth]
			{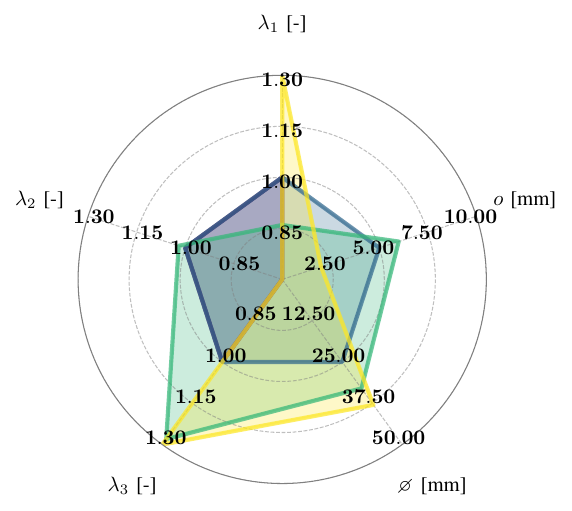}
		\end{minipage}\par
		\vspace{2mm}
		
		\includegraphics[width=\linewidth]
		{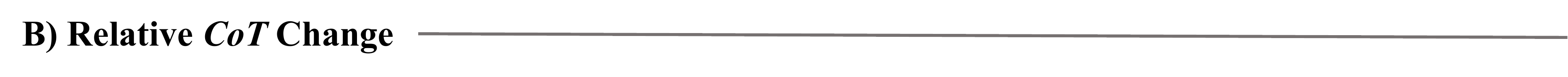}\par
		\begin{minipage}[t]{0.49\linewidth}
			\centering
			\includegraphics[width=\linewidth]
			{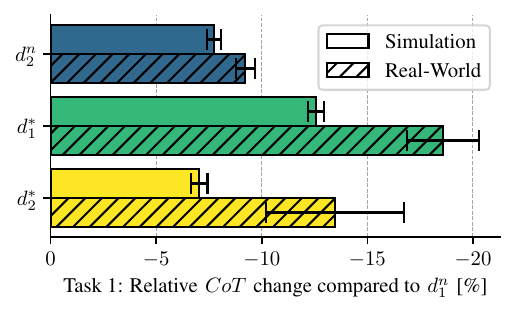}
		\end{minipage}\hfill%
		\vspace{-0mm}
		\begin{minipage}[t]{0.49\linewidth}
			\centering
			\includegraphics[width=\linewidth]
			{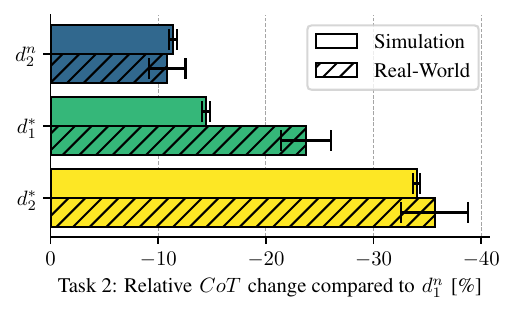}
		\end{minipage}\par
		
		\includegraphics[width=\linewidth]
		{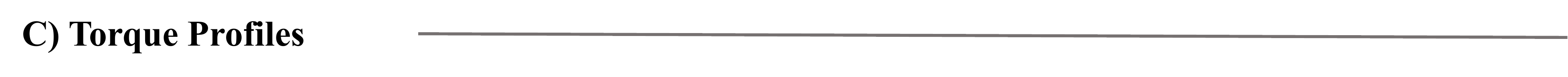}\par
		
		\includegraphics[width=\linewidth]
		{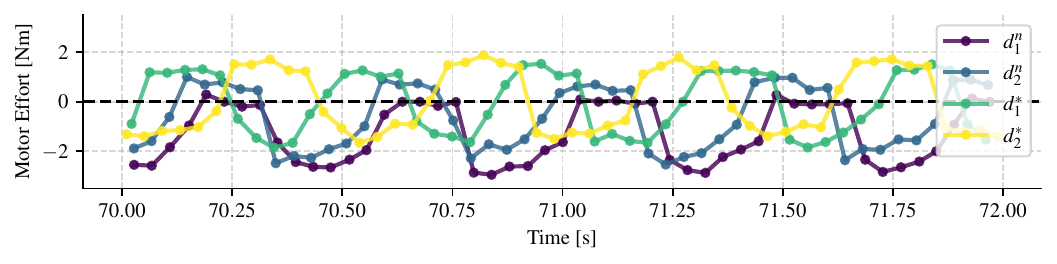}
		\vspace{-8mm}
		\caption{Task-driven co-design selects energy-efficient
			configurations.
			(A) The four designs: $d^n_1$, $d^n_2$, $d^*_1$, and $d^*_2$,
			where $d^n_1$ and $d^n_2$ share the same link parameters.
			The radar chart shows their five design parameters.
			Design colours: purple ($d^n_1$), blue ($d^n_2$),
			green ($d^*_1$), and yellow ($d^*_2$).
			(B) Relative $CoT$ changes compared with $d^n_1$
			for task 1 (left) and task 2 (right).
			Solid bars represent simulation and hatched bars represent
			hardware experiments; error bars show standard deviations.
			(C) Torque profiles of the front-right knee motor during task 1.}
		\label{fig:all}
	\end{figure}
	
	
	\subsubsection{Task 1: Efficient Inspection}
	\label{sec:task1}
	Task 1 represents an inspection along a fixed route, simplified as straight-line motion on flat ground alternating between walking and standing still (Fig.~\ref{fig:experiment2grid}A). Four repeated trials were performed for each design.
	
	The selected design for task 1, $d^*_1$, was $\lambda_1 = 0.86$, $\lambda_2 = 1.02$, $\lambda_3 = 1.28$, $\diameter = 33$\,mm, and $o = 6$\,mm, as visualized in Fig.~\ref{fig:all}(A). As shown in the left column of Fig.~\ref{fig:all}(B), this optimized design achieved a $12.6\%$ $\pm$ $0.4\%$ reduction in $CoT$ compared with ${d^n_1}$ in simulation. 
	Following zero-shot transfer, hardware experiments showed that $d^{\star}_1$ achieved an $18.6\%$ $\pm$ $1.7\%$ reduction in $CoT$ compared with ${d^n_1}$. Notably, our $CoT$ calculation includes the static power consumption of all electronics, which accounts for approximately 50\% of the total energy consumption. Because this static consumption is constant across designs and cannot be influenced by the design space, the improvement in locomotion-dependent energy use is larger than the system-level $CoT$ reduction alone suggests.
	
	\autoref{fig:all}(C) shows the front-right knee torques for task 1, revealing the effect of PEA. With the PEA disengaged (i.e., ${d^n_1}$), the knee motor exhibits the highest peak torque, and most of the knee torque lies on one side of the zero line (marked by a dotted line). In contrast, when the PEA is engaged ($d^n_2$, $d^*_1$, $d^*_2$), the torque shifts, resulting in a lower peak value and a more balanced distribution centred around zero. In particular, the optimized designs $d^*_1$ and $d^*_2$ exhibit the most balanced torque distributions and the lowest peak torques. 
	
	\subsubsection{Task 2: Efficient Payload Transport}
	\label{sec:task2}
	Task 2 represents payload transport through omnidirectional locomotion while carrying a $2$\,kg payload (approximately $30\%$ of the robot's total mass). The robot received $16$ commands spanning different walking directions and velocities over $2$ minutes, and four tests were performed for each design (Fig.~\ref{fig:experiment2grid}B).
	
	The selected design for task 2, $d^*_2$, was $\lambda_1 = 1.30$, $\lambda_2 = 0.70$, $\lambda_3 = 1.30$, $\diameter = 38$\,mm, and $o = 2$\,mm, as visualized in Fig.~\ref{fig:all}(A). The right column of Fig.~\ref{fig:all}(B) shows that, in simulation, ${d^*_2}$ achieved a $34.0\%$ $\pm$ $0.3\%$ reduction in $CoT$ relative to ${d^n_1}$ and was the most efficient among the four evaluated configurations. In the hardware experiments, ${d^*_2}$ reduced $CoT$ by $35.7\%$ $\pm$ $3.1\%$.
	
	It is worth mentioning that, although absolute $CoT$ differs between simulation and hardware, both the simulation and hardware experiments recognize the optimized morphology and the corresponding energetic benefits given the distinct task requirements, as evidenced by Fig.~\ref{fig:experiment2grid}B. 
	
	\subsection{Cross-Task Validation of Selected Co-designs}
	To assess the task specificity of the selected designs, we further evaluated ${d^*_2}$ in task 1 and ${d^*_1}$ in task 2.
	
	The left column of Fig.~\ref{fig:all}(B) shows that ${d^*_2}$ achieved a $7.0\%$ $\pm$ $0.4\%$ reduction in $CoT$ relative to ${d^n_1}$ in simulation and a $13.5\%$ $\pm$ $3.3\%$ reduction on hardware for task 1 (yellow bars). Nevertheless, ${d^*_2}$ was less energy-efficient than the task-specific design ${d^*_1}$. Conversely, ${d^*_1}$ reduced $CoT$ relative to ${d^n_1}$ by $14.4\%$ $\pm$ $0.4\%$ in simulation and $23.7\%$ $\pm$ $2.3\%$ in hardware experiments for task 2 (green bars in the right column). However, ${d^*_1}$ was less energy-efficient than the task-specific design ${d^*_2}$.
	
	These cross-validation results show that, although the optimized designs perform well when evaluated on the other task, they do not match the performance of the designs specifically optimized for their respective tasks. This indicates that the co-design process captures task-dependent optima rather than a single universally optimal solution. At the same time, the relatively strong cross-task performance suggests that these designs exploit underlying mechanical trade-offs that are partially shared across tasks, leading to overlapping regions of high-performing configurations in the design space. Consequently, the proposed pipeline does not simply identify a single best morphology but instead discovers task-aligned solutions within a structured design landscape in which multiple configurations can achieve competitive performance, depending on the task context.
	
	\subsection{Full Geometry--Compliance Co-design versus Restricted Design Scopes}

	\begin{figure}
		\centering
		
		
		\begin{minipage}{0.48\textwidth}
			\centering
			\includegraphics[width=1.0\textwidth]{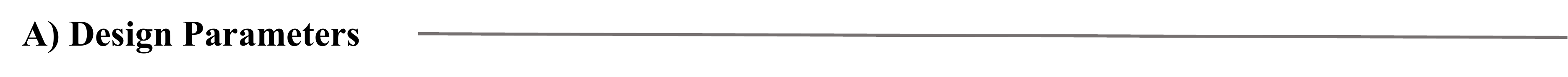}
		\end{minipage}
		
		\begin{minipage}[c]{0.48\textwidth}
			\centering
			\includegraphics[width=1.0\textwidth]{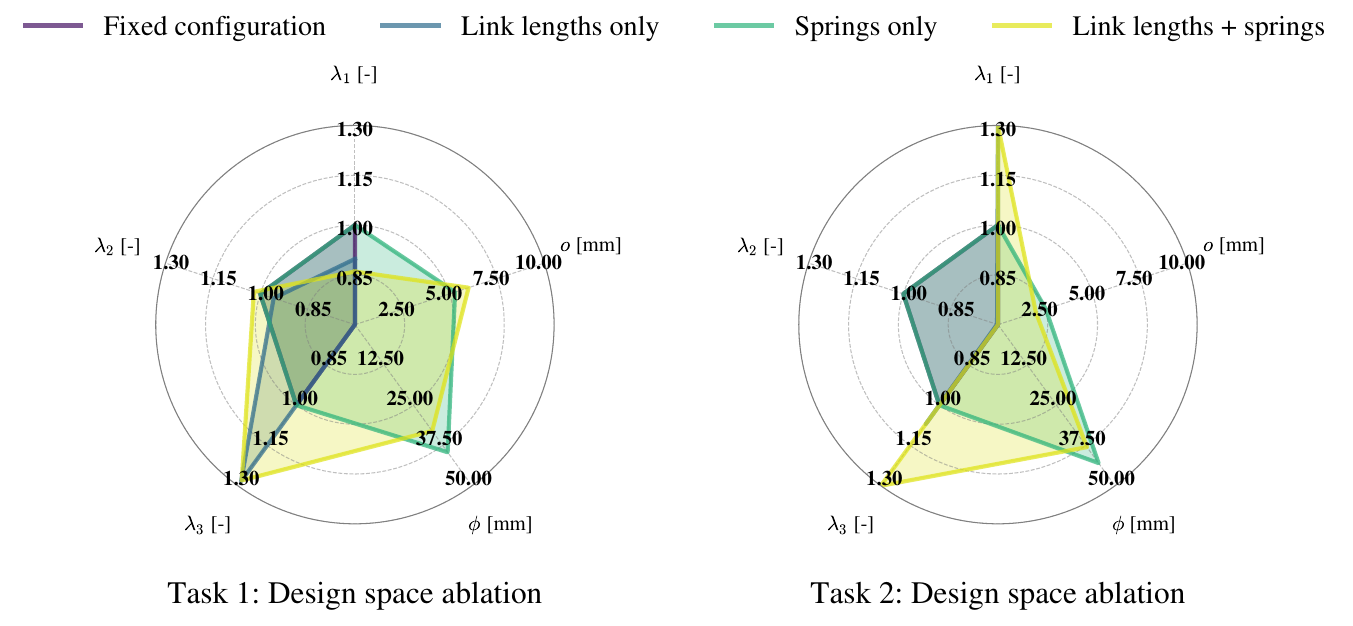}
		\end{minipage}%
		
		\vspace{-2mm}
		\begin{minipage}{0.48\textwidth}
			\centering
			\includegraphics[width=1.0\textwidth]{figures/fig_all_med.pdf}
		\end{minipage}
		
		\begin{minipage}[c]{0.48\textwidth}
			\centering
			\includegraphics[width=1.0\textwidth]{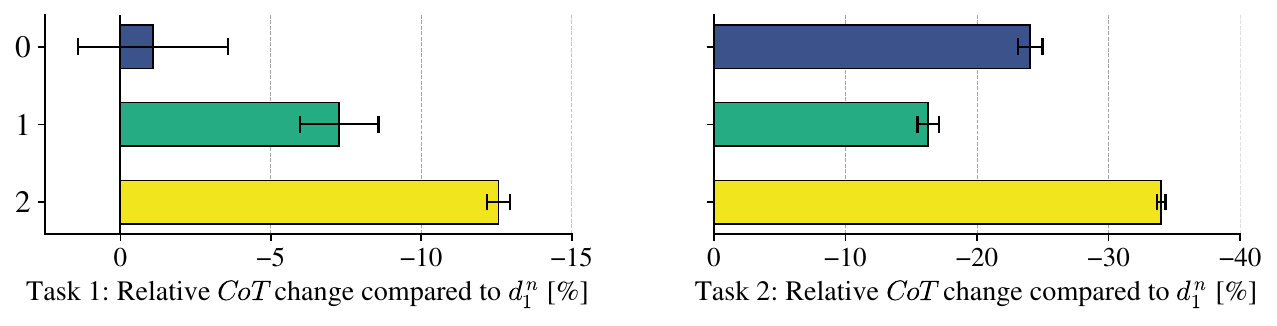}
		\end{minipage}%
		\hfill
		\caption{\textbf{Design-space ablation for tasks 1 and 2.} The left and right columns correspond to tasks 1 and 2, respectively. The top row shows the selected design parameters for the fixed, link-length-only, spring-only and full link-length--spring design scopes. The bottom row shows the relative $CoT$ changes compared with the fixed configuration, $d^n_1$; error bars indicate standard deviations. In (B), purple, green and yellow bars represent link-length-only, spring-only and full geometry–compliance co-design, respectively.}
		\label{fig:design_ablations}
	\end{figure}
	
	To isolate the contributions of geometry and passive compliance, we compared four design scopes in simulation: a fixed configuration, link length-only co-design, spring-only co-design and full link length--spring co-design. The restricted scopes optimized either the three segment lengths with the springs disengaged, or the passive knee-spring parameters with the segment lengths fixed at their nominal values, whereas the full scope optimizes all five parameters jointly.

	As shown in Fig.~\ref{fig:design_ablations}, the relative importance of the two mechanical design groups depends on the task. For task 1, link-length-only co-design produced only a small mean $CoT$ reduction of $1.1\%$ $\pm$ $2.5\%$ relative to the fixed configuration, whereas spring-only co-design achieved a larger reduction of $7.3\%$ $\pm$ $1.3\%$. For task 2, the trend was reversed: link-length-only co-design reduced $CoT$ by $24.0\% \pm 0.9\%$, compared with $16.3\%$ $\pm$ $0.8\%$ for spring-only co-design. Thus, within the tested design space, passive compliance contributes more strongly to the inspection task, while leg geometry contributes more strongly to payload transport.
	
	Jointly optimizing geometry (link lengths) and passive compliance (spring parameters) produced the largest improvement in both tasks, reducing $CoT$ by $12.6\%$ $\pm$ $0.4\%$ for task 1 and $34.0\%$ $\pm$ $0.3\%$ for task 2. The selected full designs also combine different geometry and compliance settings for the two objectives (Fig.~\ref{fig:design_ablations}, top row). These results indicate that geometry and passive compliance provide complementary design freedoms: restricting either group excludes configurations that achieve the lowest energy cost, whereas full co-design can adapt both the leg structure and passive knee mechanics to different task demands. 

	\subsection{Comparison with Alternative Co-design Baselines}
	To assess the influence of the co-design formulation, we compared our RL+BO framework with two alternative co-design approaches: a CPG controller jointly optimized with the morphology using CMA-ES \cite{hansen2001completely} and a privileged-RL controller combined with BO \cite{bjelonic2023learning}. Each method optimized the same five physical design parameters for both tasks.
	\begin{figure}
		\centering
		
		
		\begin{minipage}{0.5\textwidth}
			\centering
			\includegraphics[width=0.99\textwidth]{figures/fig_consistency_topper.pdf}
		\end{minipage}
		
		\begin{minipage}[c]{0.5\textwidth}
			\centering
			\includegraphics[width=\textwidth]{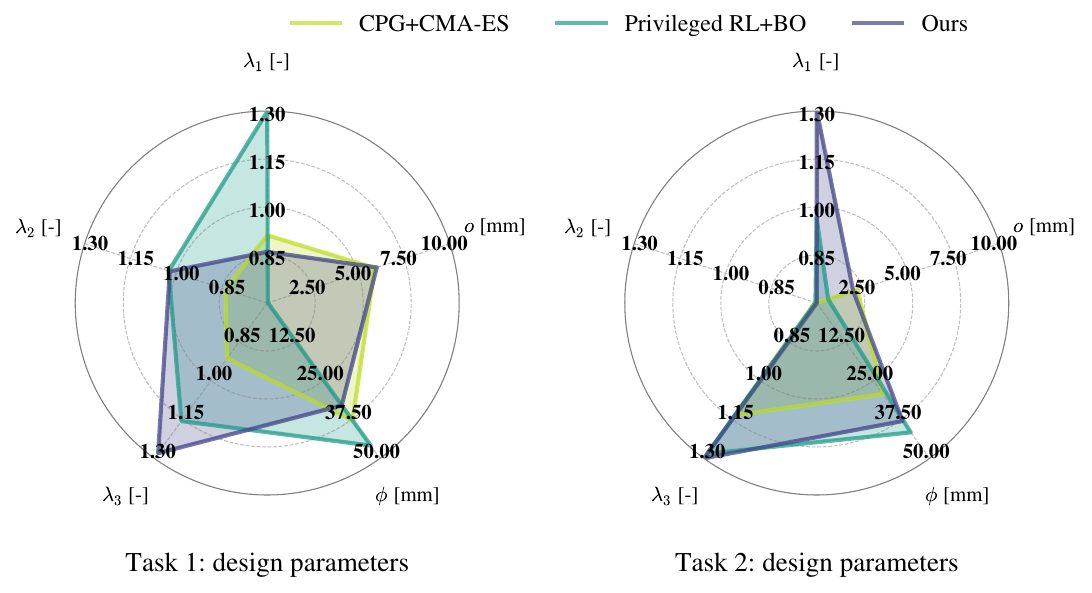}
		\end{minipage}%

		\vspace{-2mm}
		\begin{minipage}{0.5\textwidth}
			\centering
			\includegraphics[width=0.99\textwidth]{figures/fig_all_med.pdf}
		\end{minipage}
		
		\begin{minipage}[c]{0.5\textwidth}
			\centering
			\includegraphics[width=1.0\textwidth]{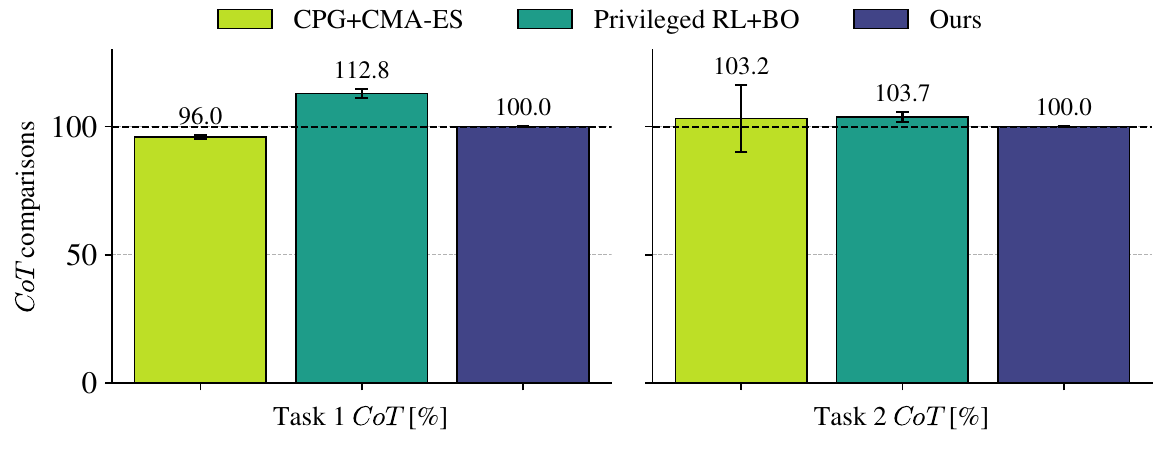}
		\end{minipage}%
		\hfill
		\caption{\textbf{Comparison with alternative co-design baselines.} The top row shows the physical configurations selected by CPG+CMA-ES, privileged RL+BO and our method. The bottom row shows the mean $CoT$ normalized to that of our method for each task; error bars indicate standard deviations.}
		\label{fig:comparevsbaselines}
	\end{figure}
	
	As illustrated in Fig.~\ref{fig:comparevsbaselines}, the three approaches selected substantially different physical configurations, indicating that the preferred morphology depends on the controller and optimization formulation. For task 1, CPG+CMA-ES achieved a mean $CoT$ of $96.0\%$ relative to our method, corresponding to a $4.0\%$ lower value, whereas privileged RL+BO reached $112.8\%$, corresponding to a $12.8\%$ higher value. Thus, the CPG-based baseline achieved the lowest mean energy cost for the inspection task.
	
	For task 2, our method achieved the lowest mean $CoT$. The normalized mean was $103.2\%$ for CPG+CMA-ES and $103.7\%$ for privileged RL+BO, corresponding to increases of $3.2\%$ and $3.7\%$, respectively, relative to our method (see the bottom right in Fig.~\ref{fig:comparevsbaselines}). The CPG+CMA-ES results also exhibited considerably greater variability for this task.
	
	Overall, these comparisons show that the proposed framework remains competitive across tasks with different locomotion demands, while the selected morphology depends on the co-design formulation.

	\subsection{Robust Locomotion on Outdoor Terrains}
	While the primary co-design optimization and quantitative evaluations were conducted on flat surfaces to ensure controlled $CoT$ comparisons, we further qualitatively assessed the robustness of the universal policy $\pi_u$ and the modular hardware on various unstructured outdoor terrains.
	
\begin{figure}[htbp]
		\centering
		\includegraphics[width=0.48\textwidth]{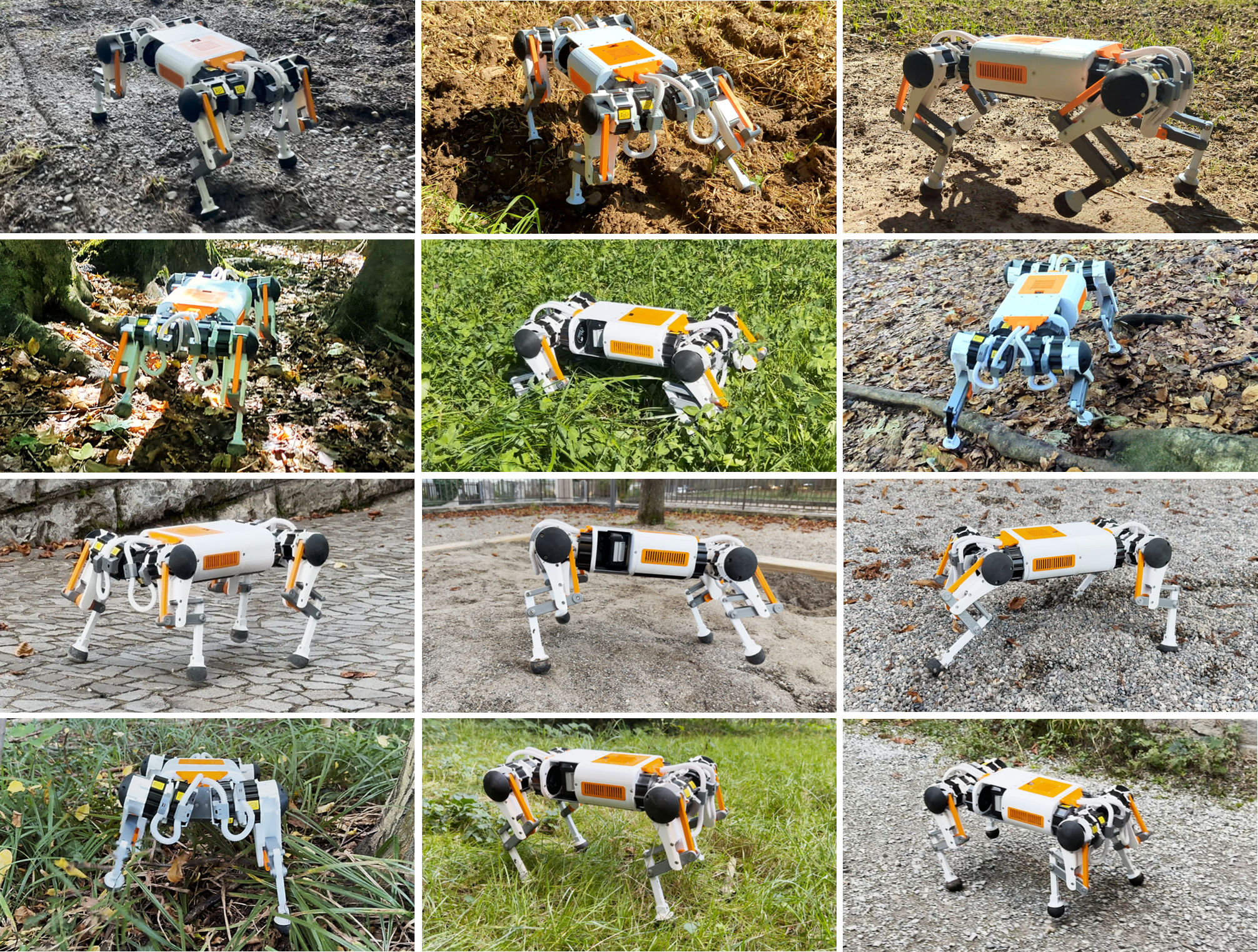}
		\vspace{-4mm}
		\caption{\textbf{Robust locomotion in outdoor environments.} Two distinct robot configurations are deployed across unstructured terrains; the configuration shown in the bottom two rows has a longer shank.}
		\label{fig:outdoor_validation}
	\end{figure}

	We deployed two distinct configurations across a range of challenging environments, including gravel, tall grass, uneven soil and forest debris (see Fig.~\ref{fig:outdoor_validation} and \href{https://youtu.be/FLeoSaKnrl8}{Movie~2}). Despite the disturbances posed by the outdoor terrain, the learning policy $\pi_u$ successfully maintained stable locomotion. 
	A separate continuous 17-min outdoor trial further demonstrated sustained locomotion without falling, controller reset or policy fine-tuning (\href{https://youtu.be/XCOe-aBKW0A}{Movie~3}).
	
	\section{Conclusion}
	This work proposes a two-step learning-based morphology-control co-design pipeline to achieve purposeful, task-driven quadrupedal design, with a bio-inspired platform. \textit{Dogger}'s three-segment legs and adjustable passive knee compliance expose complementary geometric and elastic design freedoms, while its modular construction makes these choices physically realizable. Coupled with morphology-conditioned learning, this architecture enables task-specific mechanical advantages to be identified and retained after hardware deployment. Sim-to-real deployment demonstrates that the resulting design and the control policy achieve energetic benefit satisfying different task demands.
	
	
	Using the present two-step framework, however, task-specific design selection is performed under a previously trained policy, so the selected morphology depends on the capabilities of that controller. Allowing the policy to adapt alongside morphology could reveal effective pairings that remain inaccessible to fixed-policy design search. A single-stage learning-based co-design framework could pursue this direction by jointly optimizing policy and morphology within a common learning loop. In addition, incorporating additional structural variables and practical design constraints would further strengthen the connection to robot development. Ultimately, experience accumulated across successive tasks could improve both morphology selection and policy learning, providing a basis for continual improvement of the morphology-control co-design system.

	\bibliography{sty/main.bib}
	\bibliographystyle{sty/myIEEEtran}
	
	\clearpage
	
	\appendices
	\section*{hardware platform}
	This section provides additional details on the key components of the physical robot. 
	
	\subsubsection{Data Collection Protocol}
	During hardware experiments, velocity commands are transmitted from a remote computer to the robot via a TCP connection. The commands are sent automatically according to a predefined schedule to ensure consistency across trials. The robot records sensor data at $25\,$Hz, including proprioceptive measurements, energy consumption, temperatures and control inputs. These data are then transmitted to the remote computer for post-processing. Hardware energy consumption is measured at the system level. In simulation, energy consumption is estimated using the \textit{CoT} defined in the `Methodology' Section, which is also used during the second stage of design selection.
	
	\subsubsection{Hardware Platform Overview}
	\label{sec:HW}
	To support the custom bio-inspired legs, we developed a modular robotic platform capable of accommodating different leg configurations. Existing platforms, such as the ODRI Solo 12 \cite{leziart2021implementation} and Unitree GO 1 \cite{Unitree}, present integration challenges because of physical and software constraints on incorporating new leg designs. We therefore developed a custom platform to integrate the bio-inspired legs and support the co-design pipeline.
	
	A key feature of \textit{Dogger} is that it combines off-the-shelf electronics with custom 3D-printed mechanical components made from standard materials such as PLA, PETG and TPU. These components can be fabricated using conventional FDM printers, making the platform relatively inexpensive, reproducible and adaptable to future modifications.
	
	The resulting quadruped, equipped with the nominal legs, is intermediate in size between the Unitree GO 1 and ODRI Solo 12. It weighs approximately $6.5$\,kg, with a front-to-back length of $37$\,cm and a width of $27$\,cm. The robot is actuated by Xiaomi Cybergears operating at $24$\,V and producing a peak torque of $12$\,Nm per joint. Power is supplied by a generic $2$\,Ah lithium-ion drill battery and distributed through an MJbots power-distribution board \cite{Mjbots}.
	
	An NVIDIA Jetson Orin Nano \cite{NVIDIA} is used as the onboard computer. Communication with the actuators is provided by a PEAK M.2 CAN bus card \cite{PEAK} connected through the PCIe port, with each leg assigned a separate CAN bus. Through these buses, joint position commands are transmitted and joint position estimates are retrieved. An additional CAN-FD USB dongle (CANable 2.0) \cite{canable} enables communication with the power-distribution board to monitor the robot's energy consumption. The robot's orientation is measured using an onboard MPU-6050 IMU \cite{mpu6050}, configured in digital motion processor (DMP) mode. The IMU communicates with the Jetson through the I\textsuperscript{2}C bus. 
	
	\section*{co-design pipeline settings}
	This section provides further details on the training of $\pi_u$ and the design-selection process.
	
	\subsubsection{RL Training Setup}
	We employ a universal control policy trained using proximal policy optimization (PPO)~\cite{schulman2017proximal} to generalize across a range of robot morphologies and terrains. The policy is trained in simulation with thousands of parallel environments, each assigned a different design sampled from the morphology space. In our implementation, up to 8,192 environments are used for flat-terrain training within the Brax framework~\cite{freeman2021brax}, while more complex terrain scenarios are trained using NVIDIA Isaac Lab~\cite{mittal2025isaac} with 4,096 environments. Training requires approximately 15 minutes to 2 hours on a desktop GPU (NVIDIA RTX 4070), depending on the task complexity. The resulting controller operates at a frequency of $50\,$Hz.
	
	To improve robustness in challenging environments, training incorporates a terrain curriculum and additional state information for value estimation. This setup enables stable performance across structured and unstructured terrains while maintaining a consistent control architecture across all experiments.

	\autoref{tab:training_hyperparameters} and \autoref{tab:additional_hyperparameters} list the hyperparameters used for PPO training and environment configuration, respectively. \autoref{tab:reward_functions} lists the reward components, whose symbols are defined in \autoref{tab:symbol_descriptions}. The reward weights and domain-randomization ranges are reported in \autoref{tab:reward_weights} and \autoref{tab:domain_randomization_ranges}, respectively.
	
	\subsubsection{Design Selection Setup}
	For design selection, we use the tree-structured Parzen estimator (TPE) solver \cite{ozaki2022multiobjective} provided by the Optuna library \cite{optuna_2019}. The process begins with a warm-up phase in which 100 randomly sampled designs are evaluated in parallel. Each design is evaluated for 6000 environment steps in task 1 and 15000 environment steps in task 2. 
	We perform 45 iterations with a batch size of 20 designs per iteration, requiring approximately 45 minutes of computation on the same GPU.

	\begin{table}
		\centering
		\caption{\textbf{PPO training hyperparameters.}}
		\begin{tabular}{lc}
			\hline
			\textbf{Parameter} & Value \\
			\hline
			Number of Timesteps & $150,000,000$ \\
			Episode Length & $1000$ \\
			Normalize Observations & True \\
			Unroll Length & $20$ \\
			Number of Minibatches & $32$ \\
			Number of Updates per Batch & $4$ \\
			Discounting & $0.97$ \\
			Learning Rate & $3.0 \times 10^{-4}$ \\
			Entropy Cost & $1 \times 10^{-2}$ \\
			Number of Environments & $8192$ \\
			Batch Size & $256$ \\
			Policy Hidden Layer Sizes & ($128$, $128$, $128$, $128$) \\
			\hline
		\end{tabular}
		\label{tab:training_hyperparameters}
	\end{table}
	
	\begin{table}
		\centering
		\caption{\textbf{Environment-configuration hyperparameters.}}
		\begin{tabular}{lc}
			\hline
			\textbf{Parameter} & Value \\
			\hline
			\textbf{Velocity Command Bounds} & \\
			Linear Velocity X & [$-0.6$, $0.6$] \\
			Linear Velocity Y & [$-0.6$, $0.8$] \\
			Angular Velocity Yaw & [$-1.0$, $1.0$] \\
			\hline
			\textbf{Action Space Bounds} & \\
			Lower Bounds & [$-1.0$, $0.5$, $-1.75$] \\
			Upper Bounds & [$1.0$, $2$, $0$] \\
			\hline
			\textbf{Additional Reward Parameters} & \\
			Observation Noise & $0.065$ \\
			Action Scale  & [$0.4$, $0.5$, $1.2$] \\
			Probability of Zero Command & $0.1$ \\
			\hline
		\end{tabular}
		\label{tab:additional_hyperparameters}
	\end{table}

	\begin{table*}
		\centering
		\caption{\textbf{Reward functions.}}
		\begin{tabular}{lc}
			\hline
			\textbf{Reward Component} & Expression \\
			\hline
			\textbf{Tracking Reward Functions} & \\
			Tracking Linear Velocity & 
			$\exp\left(-\frac{(v_{\text{des},x} - v_{x})^2 + (v_{\text{des},y} - v_{y})^2}{\sigma}\right)$ \\
			Tracking Angular Velocity & 
			$\exp\left(-\frac{(\omega_\text{des} - \omega_z)^2}{\sigma}\right)$ \\
			\hline
			\textbf{Regularization Reward Functions} & \\
			Linear Velocity along Z & $v_z^2$ \\
			Angular Velocity in XY & $\omega_{x}^2+\omega_{y}^2$ \\
			Orientation & $q_{\text{base},x}^2+q_{\text{base},y}^2$ \\
			Torques & $c_\text{heatloss} \cdot \sum_{i=1}^{12} \tau_i^2 + c_\text{mech\_power} \cdot \sum_{i=1}^{12} \left(\tau_i \cdot \dot q_i \right)$ \\
			Action Rate & $\sum_{i=1}^{12} \left(a_i - a_{i}^{-1}\right)^2$ \\
			Foot Air Time & $\sum_{i=1}^{4} \left((t_{i}^\text{air} - 0.1) \cdot \text{not\_contact\_filt}_i\right) \cdot \mathbb{1}(\boldsymbol{v}^{des} = 0)$ \\
			Hip Deviation from Initial Pose  & $\sum_{i=1}^{4}\left|q_{i}^\text{hip} - q^\text{default, hip}_i\right|$ \\
			Standing Deviation from Initial Pose & $\sum_{i=1}^{12} \left|q_i - q^\text{default}_i\right| \cdot \mathbb{1}(\boldsymbol{v}^{des} = 0)$ \\
			Step Height & $\sum_{i=1}^{4} \left(\text{step\_height}_i - \text{step\_height\_threshold}_i\right)^2 \cdot \text{contact\_filt}_i$ \\
			Termination & $\mathbb{1}(\text{done} \land (\text{step} < 500))$ \\
			Foot Slip & $\sum_{i=1}^{4} \left((v^{\text{foot},x,y}_i)^2 \cdot \text{contact\_filt}_i\right)$ \\
			\hline
		\end{tabular}
		\label{tab:reward_functions}
	\end{table*}
	
	\begin{table*}
		\centering
		\caption{\textbf{Symbols used in the reward functions.}}
		\begin{tabular}{ll}
			\hline
			\textbf{Symbol} & Description \\
			\hline
			$\boldsymbol{q}$ & Joint position vector for all actuators.\\
			$\boldsymbol{q}^\text{default}$ & Default joint position vector for all actuators. \\
			$\dot{q}_i$ & Angular velocity of the $i$-th actuator. \\
			$\tau_i$ & Torque applied to the $i$-th actuator. \\
			$a_i$ & Action command for the $i$-th actuator. \\
			$a_i^{-1}$ & Previous action command for the $i$-th actuator. \\
			$\boldsymbol{v}$ & Actual linear velocity vector.\\
			$\boldsymbol{v}_\text{des}$ & Desired linear velocity vector. \\
			$\boldsymbol{\omega}$ & Actual angular velocity vector. \\
			$\omega_\text{des}$ & Desired angular velocity around the $z$ axis. \\
			$\boldsymbol{q}_\text{base}$ & Orientation quaternion components representing pitch and roll of the base. \\
			$t_i^\text{air}$ & Air time of the $i$-th foot. \\
			$\text{not\_contact\_filt}_i$ & Boolean indicating whether the $i$-th foot is not in contact with the ground. \\
			$\text{contact\_filt}_i$ & Boolean indicating whether the $i$-th foot is in contact with the ground. \\
			$\boldsymbol{v}^\text{foot}_i$ & Linear velocity vector of the $i$-th foot in the $x$ and $y$ directions. \\
			$\text{step\_height}_i$ & Step height of the $i$-th foot. \\
			$\text{step\_height\_threshold}_i$ & Step-height threshold for the $i$-th foot. \\
			$c_\text{heatloss}$ & Heat loss constant. \\
			$c_\text{mech\_power}$ & Mechanical power constant. \\
			$\mathbb{1}(\cdot)$ & Indicator function (1 if the condition inside is true, 0 otherwise). \\
			$\text{done}$ & Boolean indicating if the episode has terminated. \\
			$\text{step}$ & Current simulation-step index. \\
			$\sigma$ & Variance parameter controlling the sensitivity of the exponential function. \\
			\hline
		\end{tabular}
		\label{tab:symbol_descriptions}
	\end{table*}
	
	\begin{table}
		\centering
		\caption{\textbf{Reward weights.}}
		\begin{tabular}{lc}
			\hline
			\textbf{Reward Component} & Weight \\
			\hline
			\textbf{Tracking Reward Weights} & \\
			Tracking Linear Velocity & $1.5$ \\
			Tracking Angular Velocity & $0.8$ \\
			Tracking Sigma & $0.25$ \\
			\hline
			\textbf{Regularization Reward Weights} & \\
			Linear Velocity along Z & $-2.0$ \\
			Angular Velocity in XY & $-0.05$ \\
			Orientation & $-35.0$ \\
			Torques & $-0.0015$ \\
			Action Rate & $-0.1$ \\
			Foot Air Time & $10.0$ \\
			Hip Deviation from Initial Pose & $-0.20$ \\
			Standing Deviation from Initial Pose & $-0.15$ \\
			Step Height & $-1.75$ \\
			Termination & $-1.0$ \\
			Foot Slip & $-0.1$ \\
			\hline
		\end{tabular}
		\label{tab:reward_weights}
	\end{table}

	\begin{table}
		\centering
		\caption{\textbf{Variable ranges for domain randomization.}}
		\begin{tabular}{lccc}
			\hline
			\textbf{Parameter} & Range & Unit & Notes \\
			\hline
			$\lambda_1$ & [$0.7$, $1.3$] & [-] & Factor \\
			$\lambda_2$ & [$0.7$, $1.3$] & [-] & Factor \\
			$\lambda_3$ & [$0.7$, $1.3$] & [-] & Factor \\
			$\diameter$ & [$0$, $50$] & [mm] & - \\
			$o$ & [$0$, $10$] & [mm] & - \\
			Friction & [$1.0$, $2.0$] & [-] & Multiplicative factor \\
			Position gain (Kp) & [$-2.5$, $2.5$] & [-] & Additive offset \\
			Derivative gain (Kd) & [$-0.2$, $0.35$] & [-] & Additive offset \\
			Joint damping & [$0.0$, $0.3$] & [-] & Additive offset \\
			Torso mass & [$0.75$, $1.5$] & [-] & Multiplicative factor \\
			Torso inertia & [$0.5$, $2.5$] & [-] & Multiplicative factor \\
			Torso COM position & [$0.035$, $0.05$] & [m] & Additive offset \\
			\hline
		\end{tabular}
		\label{tab:domain_randomization_ranges}
	\end{table}

	\balance

\end{document}